\documentclass[10pt,twocolumn,letterpaper]{article}

\usepackage{iccv}
\usepackage{times}
\usepackage{epsfig}
\usepackage{graphicx}
\usepackage{amsmath}
\usepackage{amssymb}

\usepackage{microtype}
\usepackage{amsfonts}
\usepackage{nicefrac}
\usepackage{booktabs}
\usepackage{color}
\usepackage{bbm}
\usepackage{pifont}%
\newcommand{\cmark}{\ding{51}}%
\newcommand{\xmark}{\ding{55}}%

\usepackage{wrapfig}

\usepackage{subcaption}

\usepackage[pagebackref=true,breaklinks=true,letterpaper=true,colorlinks,bookmarks=false]{hyperref}

\usepackage{cleveref}

\definecolor{myGreen}{rgb}{0, .6, .0}
\hypersetup{colorlinks,breaklinks,anchorcolor=darkblue,citecolor=myGreen}

\iccvfinalcopy %

\def\iccvPaperID{2742} %

\ificcvfinal\fi

\begin{document}

\title{Joint Representation Learning and Novel Category Discovery on Single- and Multi-modal Data}

\author{Xuhui Jia$^1$ \quad Kai Han$^{1,2, 3}$\thanks{~Corresponding author.} \quad Yukun Zhu$^1$ \quad Bradley Green$^1$ \vspace{0.3em}\\
$^1$Google \quad $^2$University of Bristol \quad $^3$The University of Hong Kong \\
{\tt\small \{xhjia, kaihanx, yukun, brg\}@google.com}
}

\maketitle
\ificcvfinal\thispagestyle{empty}\fi

\begin{abstract}
This paper studies the problem of novel category discovery on single- and multi-modal data with labels from different but relevant categories. We present a generic, end-to-end framework to jointly learn a reliable representation and assign clusters to unlabelled data. To avoid over-fitting the learnt embedding to labelled data, we take inspiration from self-supervised representation learning by noise-contrastive estimation and extend it to jointly handle labelled and unlabelled data. 
In particular, we propose using category discrimination on labelled data and cross-modal discrimination on multi-modal data to augment instance discrimination used in conventional contrastive learning approaches. 
We further employ Winner-Take-All (WTA) hashing algorithm on the shared representation space to generate pairwise pseudo labels for unlabelled data to better predict cluster assignments.
We thoroughly evaluate our framework on large-scale multi-modal video benchmarks Kinetics-400 and VGG-Sound, and image benchmarks CIFAR10, CIFAR100 and ImageNet, obtaining state-of-the-art results.
\end{abstract}

\section{Introduction}
With the tremendous advances in deep learning, recent machine learning models have shown superior performance on many tasks, such as image recognition~\cite{deng2009imagenet,kuznetsova2020open}, object detection~\cite{zhou2019objects,tan2020efficientdet}, image segmentation~\cite{cheng2019panoptic}, etc. While the state-of-the-art models might even outperform human in these tasks, the success of these models heavily relies on the huge amount of data with human annotations under the closed-world assumption. Applying deep learning in real (open) world brings many new challenges: it is cost-inhibitive to identify and annotate all categories, and new categories could keep emerging. Conventional methods struggle on handling unlabelled data from new categories~\cite{fontanel2020boosting}. On the flip side, real world provides rich unlabeled data, which are often multi-modal (e.g., video and audio),  allowing more possibilities for machine learning models to learn in a similar way as human. Indeed, humans learn from multi-modal data everyday with text, videos, audios, etc.

In this paper, we focus on automatically learning to discover  new categories in the open world setting. Similar to recent works ~\cite{han2019learning,han20automatically} which transfer knowledge from labelled images of a few classes to other unlabelled image collections, we formulate the problem as partitioning unlabelled data from unknown categories into proper semantic groups, while some labelled data from other categories are available. This is a more realistic setting than pure unsupervised clustering which may produce equally valid data partitions following different unconstrained criteria (e.g., images can be clustered by texture, color, illumination, etc) and closed-world recognition which can not handle unlabelled data from new categories without any labels. Meanwhile, our setting is more similar to the human cognition process where humans can easily learn the concept of a new object by transferring knowledge from known objects.

Specifically, we introduce a flexible end-to-end framework to discover categories in unlabelled data, with the goal of utilizing both labelled and unlabelled data to build unbiased feature representation, while transferring more knowledge from labelled to unlabelled data. In particular, 
we extend the conventional contrastive learning~\cite{chen2020simple,he2020moco} to consider both instance discrimination and category discrimination to learn a reliable feature representation on labelled and unlabelled data.
We also demonstrate that the cross-modal discrimination would further benefit representation learning on data with multi-modalities.
To leverage more of unlabelled data, we employ the Winner-Take-All (WTA) hashing~\cite{Yagnik2011thepower} on the shared representation space to generate pair-wise pseudo labels on-the-fly, which is the key for robust knowledge transfer from the labelled data to unlabelled data. With the weak pseudo labels, the model can be trained with a simple binary cross-entropy loss on the unlabelled data together with the standard cross-entropy loss on the labelled data. This way our model can simultaneously learn feature representation and perform the cluster assignment using an unified loss function.

The main contributions of the paper can be summarized as follows:
(1) we propose a generic, end-to-end framework for novel category discovery that can be trained jointly on labelled and unlabelled data;
(2) to the best of our knowledge, we are the first to extend contrastive learning in novel category discovery task by category discrimination on labelled data and cross-modal discrimination on multi-modal data;
(3) we propose a strategy to employ WTA hashing on the shared representation space of both labelled and unlabelled data to generate additional (pseudo) supervision on unlabeled data;
and (4) we thoroughly evaluate our end-to-end framework on challenging large scale multi-modal video benchmarks and single-modal image benchmarks, outperforming existing methods by a significant margin.

\section{Related work}
Our method is related to self-supervised learning, semi-supervised learning, and clustering, while different from each of them. We review the most relevant works below.

Self-supervised learning aims at learning reliable feature representations using the data itself to provide supervision signals during training. Many pretext tasks (e.g., relative position~\cite{doersch2015unsupervised}, colorization~\cite{zhang2017split}, rotation prediction~\cite{gidaris2018unsupervised}) have been proposed for self-supervised learning, showing promising results. Recently, the constrastive learning based methods, such as ~\cite{he2020moco} and~\cite{chen2020simple}, have attracted lots of attention by its simplicity and effectiveness. The key idea of contrastive learning is \emph{instance discrimination}, i.e., pulling similar pairs close and pushing dissimilar pairs away in the feature space. 
\cite{khosla2020supervised} studied the supervised contrastive learning on labelled data as an alternative of cross-entropy. With the labels, more positive pairs can be generated from the intra-class instances, enabling \emph{category discrimination}.
Noise-Contrastive Estimation (NCE)~\cite{gutmann2010noise,Oord2018RepresentationLW} is an effective contrastive loss widely used in these methods. 
When handling multi-modal data like videos, different self-supervised learning methods have been proposed to exploit data of different modalities, such as~\cite{patrick2020multi,alayrac20selfsup,asano2020labelling,alwassel2019self,morgado2020avid}. Among them, \cite{morgado2020avid} suggests that cross-modal discrimination can be adopted to improve the representation learning for downstream tasks like image recognition and object detection, which implies that good representations are shared between multiple views of the world. ~\cite{tian2019contrastive} shows that cross-view prediction outperforms conventional alternatives in contrastive learning on images, depth, video and flow, and more views can lead to better representation. In this paper, we consider the visual and audio modalities for cross-modal learning in videos, and present a new way to incorporate contrastive learning for both labelled and unlabelled data to bootstrap representation learning for novel category discovery. 

Semi-supervised learning~\cite{chapelle2006semi} considers the setting with labelled and unlabelled data. Specifically, the unlabelled data are assumed to come from the same classes as the labelled data. The objective is to learn a robust model making use of both labelled and unlabelled data to avoid over-fitting to the labelled data. While this problem is well studied in the literature (e.g.,~\cite{oliver2018realistic,tarvainen2017mean,rebuffi20SSL}), existing methods can not handle unlabelled data from new classes. In contrast, our method is designed to discover new categories in the unlabelled data automatically.

Clustering, which aims at automatically partitioning the unlabelled data into different groups,  has long been studied in the machine learning community. There are many classic methods (e.g.,  $k$-means~\cite{MackQueen67_Kmeans}, mean-shift~\cite{Comaniciu02meanshift}) and deep learning based methods (e.g., ~\cite{Xie16_DEC,Dizaji2017deepclustering,rebuffi20lsdc}) showing promising results. However, the definition of a cluster can be intrinsically ambiguous, because different criteria can be used to cluster the data. For example, objects can be clustered by color, shape or texture, and the clustering results will be different by taking different criteria, while these criteria cannot be predefined in the clustering methods. In this paper, we aim to learn these criteria implicitly from the labelled data and transfer them to the unlabelled data on-the-fly.

Until recently, the problem of discovering new categories in unlabelled data by exploiting the labelled data starts to draw attention. Particularly, the task of novel category discovery is formalized by~\cite{han2019learning,han20automatically,han21autonovel}.
\cite{han2019learning} introduces a method to first pretrain the model on labelled data followed by fine-tuning with an unsupervised clustering loss.
\cite{han20automatically,han21autonovel} present a three-stage method, namely, self-supervised pretraining~\cite{gidaris2018unsupervised}, supervised fine-tuning, and joint learning to transfer knowledge. 
\cite{Hsu18_L2C} and~\cite{Hsu19_MCL}, which are designed for general cross-task transfer learning, can also be applied to discover new categories in image datasets. 
These two methods require maintaining two models separately. One for binary label prediction and another for clustering.
To our knowledge, the most relevant work to ours is \cite{han20automatically}, while our method significantly differs from it in several aspects. First, our method is an end-to-end trainable framework that can be easily adopted for practical use, while \cite{han20automatically} requires three training stages, thus hampering their practical utilization; second, we conduct self-supervised learning jointly with new category discovery to bootstrap the semantic feature embedding that are more suitable for discovering new classes, rather than simply using self-supervision to provide only robust low-level features as in~\cite{han20automatically}. Third, we employ multiple random partial rankings for holistic comparison of two data points, rather than simply comparing the global ranking, which tends to be more vulnerable to noise. Last, our model can be flexibly applied on both single- and multi-modal data showing superior results, while ~\cite{han20automatically} is only applicable to images.
As will be shown in the experiments, our method substantially outperforms others in all benchmarks.
Concurrent to our work, several other methods are also proposed for novel category discovery on images showing promising performance by augmenting the training samples by mixing the labelled and unlabelled data~\cite{zhong2021openmix}, aggregating pseudo-positive pairs with contrastive learning~\cite{Zhong_2021_CVPR}, designing a unified objective~\cite{Fini_2021_ICCV}, and leveraging local part information~\cite{zhao21novel}.
\section{Method}
\label{sec:method}
Given an unlabelled collection of instances $x_{i}^{u} \in D^{u}$, our objective is to automatically partition these instances into $C^{u}$ different semantic groups. We also assume that there is a labelled collection of instances $\left(x_{i}^{l}, y_{i}^{l}\right) \in D^{l}$ where $y_{i}^{l} \in\left\{1, \ldots, C^{l}\right\}$, from which we want to transfer the knowledge to the unlabelled data so that the unlabelled data can be grouped into proper classes, while the classes in $D^{l}$ and $D^{u}$ are different but relevant.
Each $x_{i} \in D^u \cup D^l$ can be either an image or a multi-modal video consisting of a visual stream $x^v_i$ and the corresponding audio stream $x^a_i$. 

\begin{figure*}
\centering
\includegraphics[width=\linewidth]{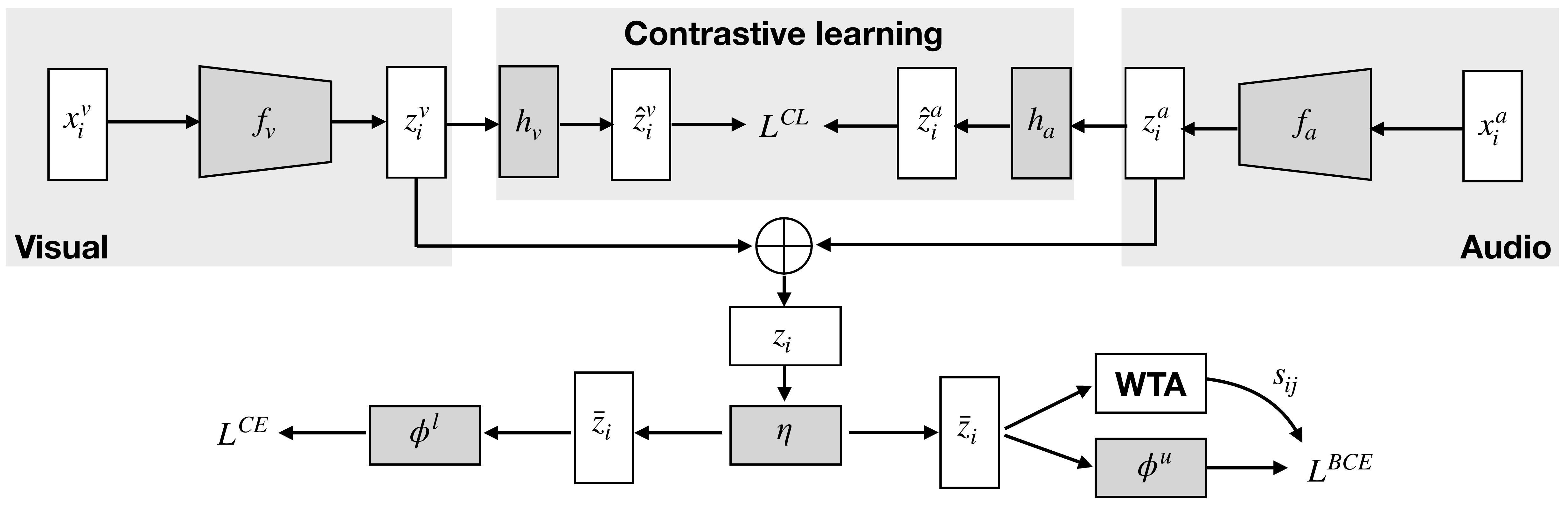}
\caption{\textbf{Overview of our end-to-end framework}. For multi-modal videos, our framework consists of two feature encoders $f_v$ and $f_a$, two MLPs $h_v$ and $h_a$ for contrastive learning, one fusion layer $\eta$, and two linear heads $\phi^l$ and $\phi^u$ for classification and clustering. The training signal for $\phi^u$ is obtained by WTA on-the-fly. 
For single-modal images, the audio encoder $f_a$ is omitted and $\eta$ turns to be an identity mapping function.
}\label{fig:splash}
\end{figure*}

Our approach is an end-to-end trainable framework that can 
jointly learn the representation and clustering assignment for the unlabelled data. \Cref{fig:splash} shows the overview of our approach. Consider a video clip $x_i = (x^v_i, x^a_i)$, the visual encoder $f_v$ and the audio encoder $f_a$ first encode the visual and audio streams into two feature vectors $z_i^{v}$ and $z_i^{a}$, which are then concatenated to form the global representation $z_i = [z_i^{v}, z_i^{a}]$ for the input video clip. A projection function $\eta$ then project $z_i$ to $\bar{z}_i$ to fuse multi-modal feature in a compact representation space. Note that $\eta$ is an identity mapping function for single-modal data. If $x_i$ is from labelled data, $\bar{z}_i$ will be sent to the linear head $\phi^l$ for supervised learning. Otherwise, it will be sent to the linear head $\phi^u$ for new category discovery.
In addition, $z_i^{v}$ and $z_i^{a}$ are further encoded to $\hat{z}_i^{v}$ and $\hat{z}_i^{a}$ by $h_v$ and $h_a$ respectively for constrastive learning to bootstrap the representation learning on both labelled and unlabelled data. $h_v$ and $h_a$ are two MLP functions followed by $\ell_{2}$ normalization, which is a common practice in contrastive learning. 

The end-to-end training of the model consists of several important components including training on labelled data with full supervision, constrative learning with instance and category discrimination on both labelled and unlabelled data, and training on unlabelled data with pair-wise pseudo labels transferred from the representation jointly learned by the previous two. To effectively transfer knowledge from the labelled to unlabelled data, we employ the winner-take-all (WTA) hashing algorithm on $\bar{z}_i$. Next, we will introduce our extended contrastive learning for novel category discovery by considering both instance and category discrimination, and knowledge transfer via the WTA hashing in more details.

\subsection{Unified contrastive learning on labelled and unlabelled data}
\label{sec:method:cl}
Given a batch of $N$ data points that are randomly drawn from both $D^{u}$ and $D^{l}$, i.e., $x_i \sim Unif(D^{u} \cup D^{l}$) for all the elements $x_i$ in batch $B$, our objective is to extract as much information as possible from $B$ to learn a representation that can be used to cluster unknown classes. Recent developments in contrastive learning focus on learning image representation under self-supervised scenario, where no labels are available, therefore it can be thought of as contrasting instance representations from a pair of data augmentation (positive) to those of other samples (negative). 
In our case, we have a mixed setting with both labelled and unlabelled data. We adopt the contrastive learning to jointly learn from the mixed dataset for novel category discovery. Next, we first demonstrate our unified contrastive learning on single-modal data, and then extend it to the more flexible multi-modal scenario.
\subsubsection{Single-modal learning}
Let $x_i$ be a single-modal data point and $i\in \mathcal{N}=\{1, \dots, 2N\}$ be the index of pairs of augmented samples from the batch $B$. With the embedded representation $\hat{z}_i$ of $x_i$, we adopt NCE~\cite{gutmann2010noise} as our contrastive loss function for instance discrimination, which can be written as:
\begin{equation}
\label{eq:nce_s}
\mathcal{L}^{NCE-I}_{i} = - \log \frac{\exp \left(\hat{z}_i \cdot \hat{z}_{i'} / \tau\right)}{\sum_{n}  \mathbbm{1}_{[n \ne i]} \exp \left(\hat{z}_i \cdot \hat{z}_n / \tau\right)},
\end{equation}
where $\hat{z}_{i'}$ is the augmented counterpart of $\hat{z}_i$, $\mathbbm{1}_{[n \ne i]}$ is an indicator function evaluating 1 \emph{iff}  $n \ne i$, and $\tau$ is a scalar temperature.
This loss is widely used in conventional contrastive learning approaches.

In our case, some data points in the mini-batch $B$ are accompanied with labels. If $x_i$ is a labelled data point in the batch, besides considering only $x_i$ and its transformed counterpart as a positive pair, other data points from the same class as $x_i$ can also be paired with $x_i$ to form more positive pairs to be pulled together, allowing category discrimination. The contrastive loss for category discrimination with these additional positive pairs can be written as:
\begin{equation}
\label{eq:nce_m}
\mathcal{L}^{NCE-C}_{i} = - \frac{1}{|Q(i)|}  \sum_{q \in Q(i)}\log \frac{\exp \left(\hat{z}_i \cdot \hat{z}_{q} / \tau\right)}{\sum_{n} \mathbbm{1}_{[n \ne i]} \exp \left(\hat{z}_i \cdot \hat{z}_n / \tau\right)}, 
\end{equation}
where $Q(i)=\{q \in \mathcal{N} \setminus i: y_q = y_i\}$ denotes the indices of other data points which have the same label as $x_i$ in the batch $B$. Note that this is effective only for labelled data, as $Q(i)=\emptyset$ for unlabelled data. 
Therefore, 
the unified contrastive loss can be written as:
\begin{equation}
\mathcal{L}^{CL} = \frac{1}{2N} \sum_{i}^{2N} \left( \mathcal{L}^{NCE-I}_{i} + \mathcal{L}^{NCE-C}_{i} \right) .
\end{equation}

Note that there is only one positive pair in $\mathcal{L}^{NCE-I}_{i}$ which aims at instance discrimination, whereas multiple other samples from same class are considered as positive in $\mathcal{L}^{NCE-C}_{i}$ which aims at category discrimination. 
$\mathcal{L}^{NCE-I}_{i}$ and $\mathcal{L}^{NCE-C}_{i}$  have exactly the same denominator, which is the summation of $2N-1$ scores for both positive and negative pairs.
In this way, both labelled and unlabelled data are used for representation learning while making full use of the labels contained in the labelled data. 
$\mathcal{L}^{NCE-C}_{i}$ is a critical complementary to $\mathcal{L}^{NCE-I}_{i}$, resulting in a more discriminative representation space for robust clustering as will be seen in the experiments.

\subsubsection{Multi-modal learning}

For multi-modal data like videos, besides the conventional within-modal contrastive learning for single-modal data like images, we can also have the additional cross-modal option. As noted by~\cite{alwassel2019self,morgado2020avid}, cross-modal agreement leads to better representation than within-modal agreement in self-supervised representation learning for supervised downstream tasks like object recognition and detection. However, our setting contains both labelled data and unlabelled data, resulting in a mixture of instance discrimination and category discrimination. It is not immediately obvious whether the within-modal or cross-modal choice is more effective under such a setting. For within-modal case, we can either discriminate visual or audio samples. 
Let the embedded representation for a multi-modal data point $x_i$ be
$\hat{z}_i = \{\hat{z}_i^v, \hat{z}_i^a \}$. We define modality selecting functions $g_0$ and $g_1$ to allow either within-modal discrimination (e.g., $g_0(\hat{z}_i) = \hat{z}_i^v$ and $g_1(\hat{z}_i) = \hat{z}_i^v$) or cross-modal discrimination (e.g., $g_0(\hat{z}_i) = \hat{z}_i^v$ and $g_1(\hat{z}_i) = \hat{z}_i^a$). The instance discrimination and category discrimination objectives  in Eq.~(\ref{eq:nce_s}) and Eq.~(\ref{eq:nce_m}) can then be rewritten as 

\begin{equation}
\label{eq:nce_s_m}
\mathcal{L}^{NCE-I}_{i} = - \log \frac{\exp \left(g_0(\hat{z}_i) \cdot g_1(\hat{z}_{i'}) / \tau\right)}{\sum_{n}  \mathbbm{1}_{[n \ne i]} \exp \left(g_0(\hat{z}_i) \cdot g_1(\hat{z}_n / \tau\right)},
\end{equation}
 and 
\begin{multline}
\label{eq:nce_m_m}
\mathcal{L}^{NCE-C}_{i} = \\
- \frac{1}{|Q(i)|}  \sum_{q \in Q(i)}\log \frac{\exp (g_0(\hat{z}_i) \cdot g_1(\hat{z}_{q}) / \tau)}{\sum_{n} \mathbbm{1}_{[n \ne i]} \exp (g_0(\hat{z}_i) \cdot g_1(\hat{z}_n) / \tau)}.
\end{multline}

As will be shown in the experiments, after investigating different strategies, we find that the cross-modal contrastive learning can produce better representation for novel category discovery.
This might be due to the fact that cross-modal representations are intrinsically different, thus reducing the chance of obtaining trivial solutions such as discriminating objects by only verifying the most salient features in visual or audio modality alone.
Therefore, unless stated otherwise, the modality selecting functions $g_0$ and $g_1$ choose visual and audio representations respectively in our formulation.

\subsection{Knowledge transfer via Winner-take-all hash}
To leverage the labelled data to help novel category discovery in unlabelled data, 
we transfer knowledge from labelled data to unlabelled data by adopting the Winner-Take-All (WTA) hash~\cite{Yagnik2011thepower} during training. WTA is a sparse embedding method that maps the feature vectors into integer codes. 
In this work, we employ WTA to measure the similarity between each pair of unlabelled data points from new categories in the shared embedding space of both labelled and unlabelled data, so that we can transfer knowledge from the labelled categories to the unlabelled ones.

\paragraph{Winner-take-all hash~~\cite{Yagnik2011thepower}}
The idea of WTA is to measure similarity between high-dimensional feature vectors by comparing multiple partial ranking statistics. The WTA algorithm works as follows. First, we randomly generate a set of $H$ permutations $\mathcal{P} = \{\rho_1, ..., \rho_H\}$. For an unlabelled sample $x_i\in D^u$, we extract its feature vector $\bar{z}_i$. 
We then apply each $\rho_h \in \mathcal{P}$ on $\bar{z}_i$ to obtain a transformed feature vector $\rho_h(\bar{z}_i)$, i.e., a shuffled version of $\bar{z}_i$. 
Let ${c}^h_i$ be the index of maximum value in the first $k$ elements of $\rho_h(\bar{z}_i)$. We can then obtain the  WTA hash code by ${c}_i = ({c}^1_i, ..., {c}^H_i)$ for each $x_i$.

\paragraph{WTA for novel category discovery}
In our case, we employ WTA hash code to measure the similarity $s_{ij}$ between $x_i$ and $x_j$ to generate pairwise pseudo labels for novel category discovery, by simply comparing their WTA hash codes ${c}_i$ and ${c}_j$:
\begin{equation}
s_{ij}=
\left\{\begin{array}{l}
1, \ \mathbf{1}^\text{T}\cdot({c}_i={c}_j) >= \mu \\
0, \ \text{otherwise}
\end{array}\right.
\label{e:wta},
\end{equation}
where $\mu$ is an empirical scalar threshold. Note that WTA is only applied during training to generate binary pseudo labels, and it is not needed at test time.

As discussed in~\cite{Yagnik2011thepower}, the precise values in the high-dimensional embedding is often not important, and the relative magnitude matters more. 
While requiring the total orderings to be identical is too strict for real application, as noise inevitably exists even using ranking statistics, therefore, WTA introduces multiple partial order statistics. 
Besides introducing more resilience to noise, we further emphasize that the partial orders for local rank correlation also captures the relative structural information of the objects. Intuitively, the global ranking statistics may only consider the most salient features in the embedding space, while the multiple partial orders spread more in the embedding space, thus capturing more structural information. As shown in~\cite{zhou20lookinto}, modern CNNs are very likely to make decisions by focusing on the salient patterns while overlooking the holistic structural composition. 
As will be seen in the experiments, WTA outperforms other alternatives for generating pairwise pseudo labels to discover new categories. 

After applying the WTA algorithm, we can obtain pairwise similarity $s_{ij}$ for each pair of unlabelled data points with Eq.~(\ref{e:wta}). Assume we have $M$ unlabelled samples in a mini-batch. By using these pairwise similarities as pseudo labels, we can then train the model with binary cross-entropy loss to simultaneously learn representation and cluster assignments on the unlabelled data from new classes:
\begin{align}
\mathcal{L}^{BCE}  =& -\frac{1}{M^{2}} \sum_{i=1}^{M} \sum_{j=1}^{M}[s_{i j} \log \phi^{u}(\bar{z}_{i})^{\top} \phi^{u}(\bar{z}_{j})+\\
&(1-s_{i j}) \log (1-\phi^{u}(\bar{z}_{i})^{\top} \phi^{u}(\bar{z}_{j}))],
\end{align}
where $\phi^{u}: \mathbb{R}^{d} \rightarrow \mathbb{R}^{C^{u}}$  is a non-linear function mapping $\bar{z}_i$ into an embedding space with the same dimension as the number of classes in the unlabelled data followed by softmax normalization. In this way, we can obtain the cluster assignment for each unlabelled sample by indexing the location of the maximum value in $\phi^{u}(\bar{z}_{i})$ after training, without requiring another offline clustering procedure for class assignment.

\subsection{Joint learning objective}
Inspired by the literature of semi-supervised learning~\cite{tarvainen2017mean}, we include a consistency regularization loss on both labelled and unlabelled data. The purpose of such a consistency loss is to enforce the class predictions on a data point $x_i$ and its transformed counterpart $x_i'$ to be the same. This is especially important for unlabelled data samples. By enforcing the consistency, $x_i$ and $x_i'$ will be treated as a positive pair regardless of the WTA hash code, as the WTA hash code of $x_i$ might be different from that of $x_i'$, thus smoothing the training. The consistency loss is commonly implemented as the mean squared error (MSE) between the class predictions. Let $\mathcal{L}^{MSE}$ be the consistency loss and $\mathcal{L}^{CE}$ be the cross-entropy loss on labelled data. 
The consistency loss between $x_i$ and its transformed version $x_i'$ is defined as
\begin{equation}
    \mathcal{L}^{MSE}_i = (\phi(\bar{z}_i) - \phi(\bar{z}_i'))^2, 
\end{equation}
where $\phi$ is $\phi^l$ or $\phi^u$ depending on the input. 
The overall training loss of our end-to-end framework can then be written as \begin{equation}
    \mathcal{L} = \mathcal{L}^{CE} + \mathcal{L}^{BCE} + (1-\omega(r))\mathcal{L}^{CL}  + \omega(r)\mathcal{L}^{MSE}, 
\end{equation}
where $\omega(r)$ is a ramp-up function slowly increasing from 0 to 1 along with the training. In our experiment, we follow~\cite{laine2016temporal} to use $\omega(r)=\lambda e^{-5\left(1-\frac{r}{T}\right)^{2}}$ where $r$, $T$ and $\lambda$ are current epoch number, total number of epochs and a positive scalar factor. We set $(1-\omega(r))$ as the weight for contrastive learning. At the early stages,  the cluster assignment predictions are noisy and we expect the model to focus more on representation learning, thus a higher weight is set to representation learning and a lower weight is set for consistency. In the late stages, the representation is good enough and we would like the model to focus on novel category discovery, therefore the weight is higher for consistency loss and lower for contrastive learning.
\section{Experiments}
\textbf{Benchmarks and evaluation metric.} We comprehensively evaluate our approach for novel category discovery on large-scale image benchmarks including ImageNet~\cite{deng2009imagenet}/CIFAR-10~\cite{Krizhevsky09cifar}/CIFAR-100~\cite{Krizhevsky09cifar} and video benchmarks including Kinetics-400~\cite{kay2017kinetics}/ VGG-Sound~\cite{chen2020vggsound}. We follow~\cite{han20automatically} to split the ImageNet/CIFAR-10/CIFAR-100 to have 30/5/20 classes in the unlabelled classes. For fair comparison with~~\cite{han2019learning,han20automatically,Hsu18_CCL} three 30-class splits are used for ImageNet, and the results are averaged over the three splits. We split Kinetics-400/VGG-Sound to have 50/39 classes in the unlabelled data, which are much more challenging than the image benchmarks. 
As only URLs are publicly available for video datasets, by the time of our experiments, we have 170k video clips with sound in Kinetics-400 and 183k videos with sound in VGG-Sound.
To measure the novel category discovery accuracy, we adopt the widely used average clustering accuracy defined as: $\max_{e\in \mathcal{P}(C^u)} = \frac{1}{U} \sum_{i=1}^{U} 1\{ {y^u_i} = e(\hat{y}^u_i)\}$, where $y^u_i$ and $\hat{y}^u_i$ denote the ground-truth label and predicted cluster assignment for each unlabelled data point, $U$ is the total number of unlabelled instances in the whole dataset, $\mathcal{P}(C^u)$ denotes the set of all possible permutations of $C^u$ elements, $e$ is an arbitrary permutation in $\mathcal{P}(C^u)$. We obtain the optimal permutation $e^*$ by Hungarian algorithm~\cite{kuhn1955hungarian}. Note that as no supervision is used for unlabelled data, the same data are used for both training and evaluation following standard practice ~\cite{ji2019invariant,han2019learning}.

\textbf{Implementation details.} 
We use R3D-18~\cite{tran2018closer,ji20123d} as the video encoder and ResNet-18~\cite{he2016deep} as the image and audio encoder. The feature vector dimension is 512 for both encoders. R3D-18 is an effective yet lightweight model for video recognition tasks, which has been shown effective for multi-modal self-supervised learning~\cite{patrick2020multi}, allowing the use of a relatively larger batch size for contrastive learning. The choice of ResNet-18 for image datasets is to follow~\cite{han2019learning,han20automatically} for fair comparison.
The MLPs for contrastive learning consist of a hidden layer of size 512, a linear layer of size 128, and an $\ell$-2 normalization layer~\cite{wu2018unsupervised}. 
The output dimension of the hidden layer $\eta$ is 512. 
On image benchmarks, we follow SimCLR~\cite{chen2020simple} to randomly apply cropping, resizing, horizontal flip, color distortion, and Gaussian blur for data augmentation. On video benchmarks, we follow~\cite{szegedy2015going} to use the input video and audio clips of 1 and 2 second duration respectively. Video frames are resized such that the shorter side has a size of 128, followed by random cropping with size 112. We preprocess the audio by randomly sampling within 0.5 second of the video and compute a log mel bank features with 257 filters and 199 time-frames, followed by SpecAugment~\cite{park2019specaugment}. During evaluation, we follow~\cite{patrick2020multi} to uniformly sample 10 clips from each video and take the mean score for prediction. 
For WTA, we set $H$ to be equal to the feature dimension (i.e., 512), and  follow~\cite{Yagnik2011thepower} to set $k$ as 4. For the threshold $\mu$, we empirically set it to 240 in our experiments.
We train our models with SGD~\cite{sutskever2013importance} and use a batch size of 1024 for all benchmarks except CIFAR-10/CIFAR-100, which we use a batch size of 256. The models are trained with $8\times8$ TPUv2 Dragonfish devices.

\subsection{Novel category discovery on image benchmarks}
In~\cref{tab:it_cmp}, we compare our approach with the $k$-means baseline and the previous state-of-the-art methods. By comparing row 10 with rows 1-6, we can clearly see that our method substantially outperforms the $k$-means baseline and existing methods. For example, our method outperforms the previous state-of-the-art method~\cite{han20automatically}, denoted as RS in row 5, by 3\%/3.2\%/4.2\% on CIFAR-10/CIFAR-100/ImageNet. We also report the incremental learning result of RS ~\cite{han20automatically} in row 6, where an additional step is introduced to create interactions between labelled and unlabelled heads.
The $k$-means baseline in row 1 is evaluated on the features of unlabelled data extracted using the model pretrained on labelled data by cross-entropy loss. We further enhance the $k$-means baseline, KCL~\cite{Hsu18_L2C} and MCL~\cite{Hsu19_MCL} by introducing the same constrative learning used in our framework (see rows 7-10). The contrastive learning successfully boosts the performance for all of them, but the results still largely lag behind our method. 
For example, after the enhancement, the ACC of KCL is increased to 73.9\%/57.4\%/ 74.3\% on CIFAR-10/CIFAR-100/ImageNet, while the results of our method are 93.4\%/76.4\%/86.7\%, demonstrating the effectiveness of our approach on image benchmarks.

\begin{table}[htb]
\centering
\caption{\textbf{Novel category discovery on image benchmarks.} ``CL'' denotes the contrastive learning we introduce in~\cref{sec:method:cl}, which performs both instance discrimination and category discrimination.  Results on CIFAR-10/CIFAR-100 are averaged over 10 runs. }
\label{tab:it_cmp}
\resizebox{0.48\textwidth}{!}{%
\begin{tabular}{l l c c c}
\toprule
 No & Method & CIFAR-10 & CIFAR-100 & ImageNet \\
\midrule
   (1) & $k$-means~\cite{MackQueen67_Kmeans}& 65.5 $\pm$ 0.0\% & 56.6 $\pm$ 1.6\% & 71.9\% \\ 
   (2) & KCL~\cite{Hsu18_L2C}& 66.5 $\pm$ 3.9\% & 14.3 $\pm$ 1.3\% & 73.8\% \\ 
   (3) & MCL~\cite{Hsu19_MCL}& 64.2 $\pm$ 0.1\% & 21.3 $\pm$ 3.4\% & 74.4\% \\ 
   (4) & DTC~\cite{han2019learning} w/ RotNet & 88.7 $\pm$ 0.3\% & 67.3 $\pm$ 1.2\% & - \\ 
   (5) & RS~\cite{han20automatically}& 90.4 $\pm$ 0.5\% & 73.2 $\pm$ 2.1\% & 82.5\% \\ 
   (6) & RS~\cite{han20automatically} w/ I.L.& 91.7 $\pm$ 0.9\% & 75.2 $\pm$ 4.2\% & - \\ 
\midrule  
   (7) & $k$-means~\cite{MackQueen67_Kmeans} w/ CL & 73.2  $\pm$ 0.1\% & 58.8 $\pm$ 1.5\% & 73.1\% \\ 
   (8) & KCL~\cite{Hsu18_L2C} w/ CL & 73.9 $\pm$ 0.1\% & 57.4 $\pm$ 3.6\% & 74.3\% \\ 
   (9) & MCL~\cite{Hsu19_MCL} w/ CL & 72.3 $\pm$ 0.2\% & 50.8 $\pm$ 3.1\% & 75.2\% \\ 
   (10) & RS~\cite{han20automatically} w/ CL &  89.0 $\pm$ 0.1\%  & 54.6 $\pm$ 1.8\% & 82.7\% \\
\midrule
   (11) & Ours & \textbf{93.4 $\pm$ 0.6\%} & \textbf{76.4 $\pm$ 2.8\%} & \textbf{86.7\%} \\
\bottomrule
\end{tabular}
}
\end{table}

\begin{figure}[h]
\centering
\begin{subfigure}{.23\textwidth}
  \centering
  \includegraphics[width=1.0\linewidth]{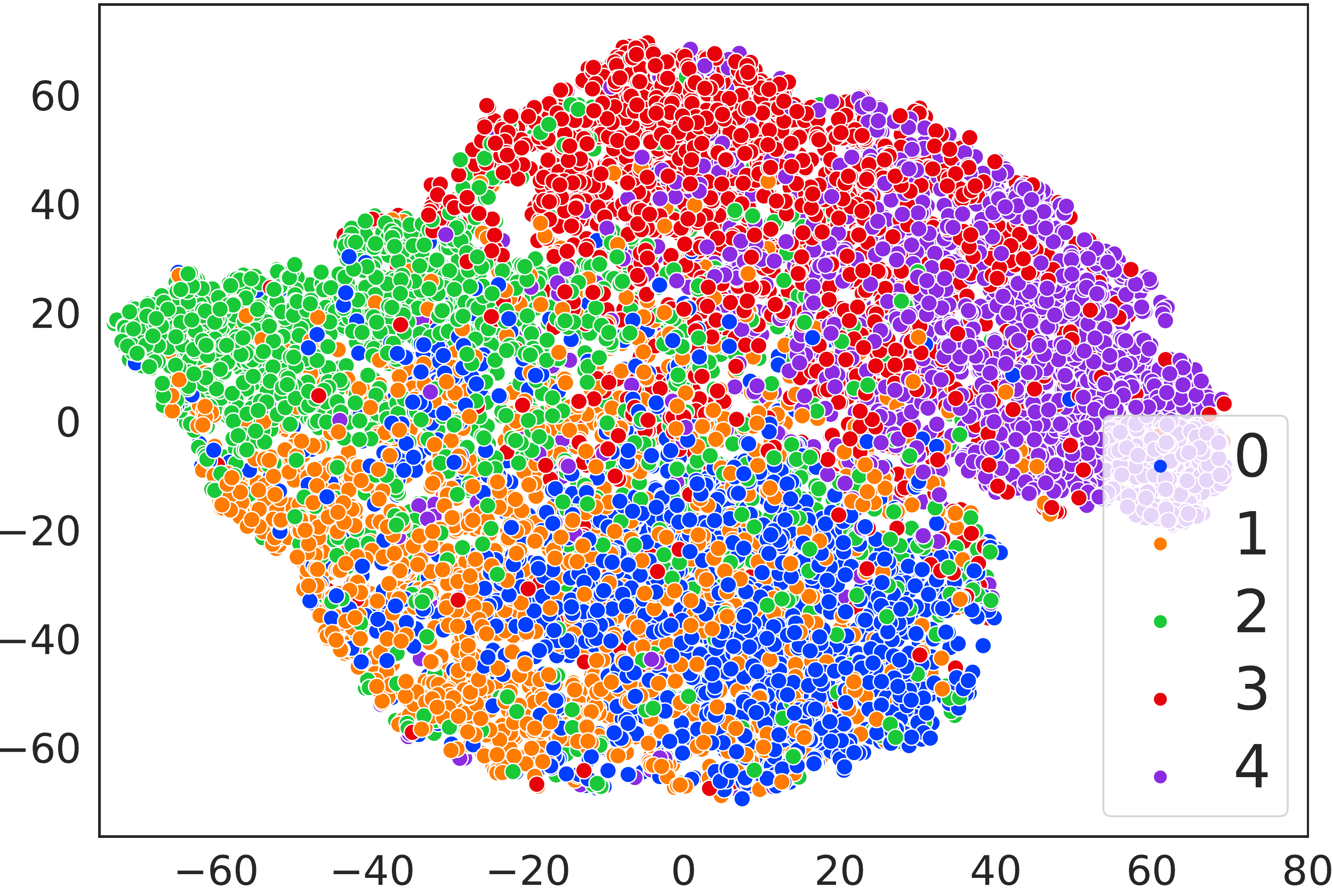}  
\end{subfigure}
\begin{subfigure}{.23\textwidth}
  \centering
  \includegraphics[width=1.0\linewidth]{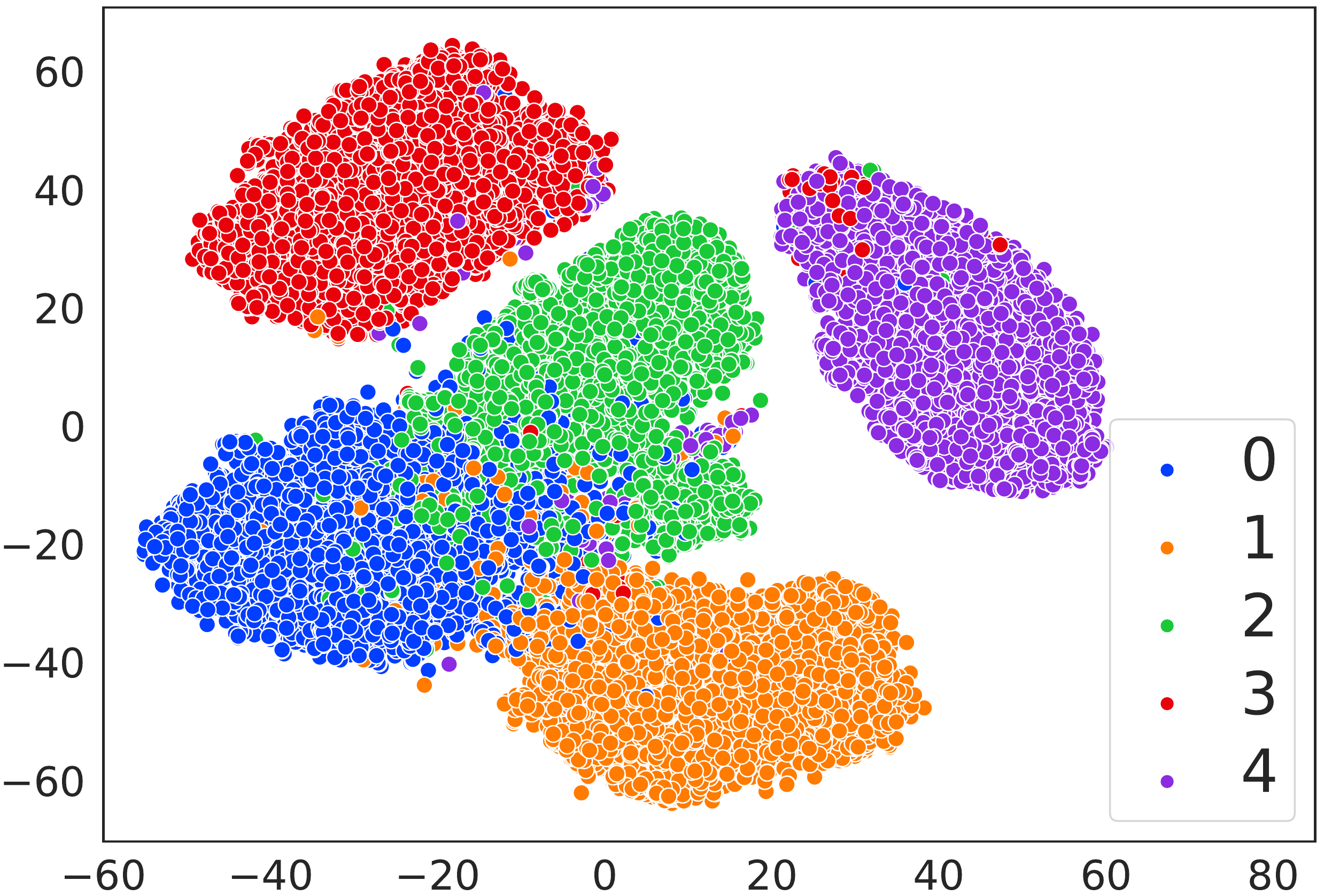}  
\end{subfigure}
\caption{\textbf{Qualitative results on unknown classes of CIFAR-10.} Left: model trained on the labelled data only. Right: our model after end-to-end training on both labelled and unlabelled data. Color denotes the ground truth.}
\label{fig:cifar_quali}
\end{figure}

In~\cref{fig:cifar_quali} we show the t-SNE projection for the features of data from the 5 unlabelled classes in CIFAR-10. The features are extracted using the model pretrained on the labelled data (\cref{fig:cifar_quali} left) and using our model (\cref{fig:cifar_quali} right) respectively. We can see that the embedding is rather cluttered using the model pretrained on the labelled data, while our model can successfully partition the unlabelled data into tight semantic groups.

\subsection{Novel category discovery on video benchmarks}

In~\cref{tab:vt_cmp}, we compare our framework with $k$-means baseline and~\cite{han20automatically} on Kinetics-400 and VGG-Sound benchmarks. In rows 1-3, we train the models on labelled data with cross-entropy loss, and extract the features on unlabelled data for $k$-means. We validate features from multi-modalities as well as each single-modality. The multi-modal features are more effective than the single-modal features. 
We also train the model with self-supervised contrastive learning and run $k$-means one the learned features (row 4). This is less effective than the baseline in row 3.
To compare with the previous state-of-the-art method~\cite{han20automatically}, which is designed for image category discovery, we implement its multi-modal counterpart (row 5) and also enhance it by replacing the original RotNet pretraining with our improved contrastive learning (row 6). 
It can be seen from row 7 that the contrastive learning introduced in our framework can further boost the performance of~\cite{han20automatically}, and our approach consistently outperforms the $k$-means baseline and  ~\cite{han20automatically} by a large margin. For example, our method achieves $56.5\%$/$50.0\%$ ACC on Kinetics-400/VGG-Sound, while~\cite{han20automatically} only gives $31.2\%$/$38.6\%$ ACC. Interestingly, we found that~\cite{han20automatically} results in row 5-6 are less effective than the $k$-means baseline. We conjecture that ranking statistics in~\cite{han20automatically} only considers the most salient features when generating pseudo labels to train on unlabelled data. This might be sufficient for images but not for complex videos, e.g., different actions may share background or sub-actions, and ranking statistics may wrongly recognize them as positive pairs, leading to poor performance. In contrast, WTA considers holistic structure, and suffers much less from this. 

\begin{table}[htb]
\centering
\caption{\textbf{Novel category discovery on video benchmarks.}}
\label{tab:vt_cmp}
\resizebox{0.45\textwidth}{!}{%
\begin{tabular}{l l c c c c}
\toprule
  No & Method & audio & video & Kinetics-400 & VGG-Sound \\
\midrule
   (1) & $k$-means~\cite{MackQueen67_Kmeans} & \xmark & \cmark & 40.9\%  & 32.4\% \\
   (2) & $k$-means~\cite{MackQueen67_Kmeans} & \cmark & \xmark & 18.7\%  & 41.7\% \\
   (3) & $k$-means~\cite{MackQueen67_Kmeans} & \cmark & \cmark & 41.1\%  & 43.4\% \\
\midrule  
   (4) & CL~\cite{chen2020simple} + $k$-means & \cmark & \cmark & 34.7\% & 28.1\% \\
\midrule  
   (5) & RS ~\cite{han20automatically} w/ RotNet & \cmark & \cmark & 31.2\%  & 38.6\% \\
   (6) & RS ~\cite{han20automatically} w/ CL & \cmark & \cmark & 33.5\%  & 42.2\% \\
\midrule
    (7) & Ours & \cmark & \cmark & \textbf{56.5\%}  & \textbf{50.0\%} \\
\bottomrule
\end{tabular}
}
\end{table}

\begin{figure}[h]
\centering
\begin{subfigure}{.23\textwidth}
  \centering
  \includegraphics[width=1.0\linewidth]{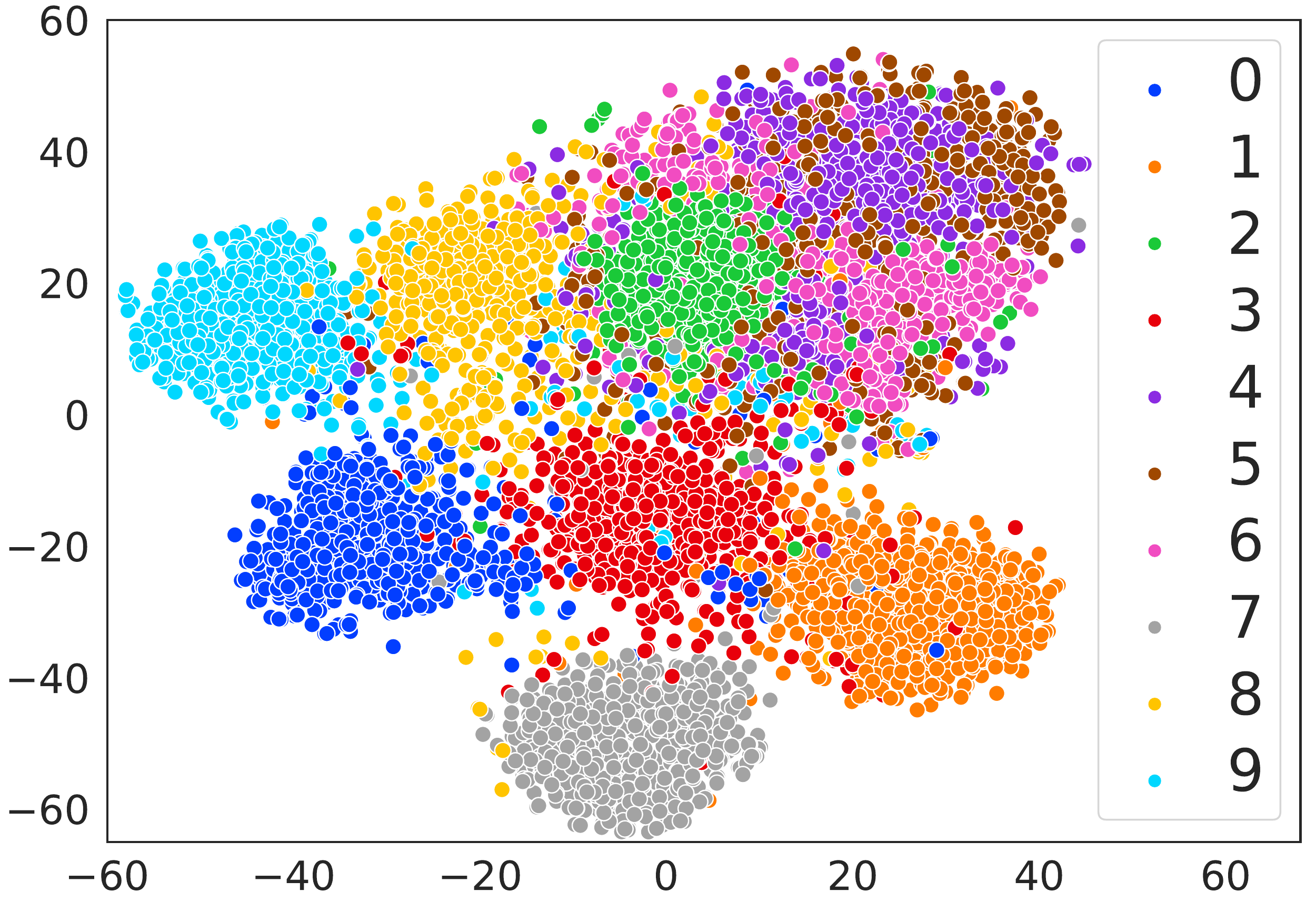}  
\end{subfigure}
\begin{subfigure}{.23\textwidth}
  \centering
  \includegraphics[width=1.0\linewidth]{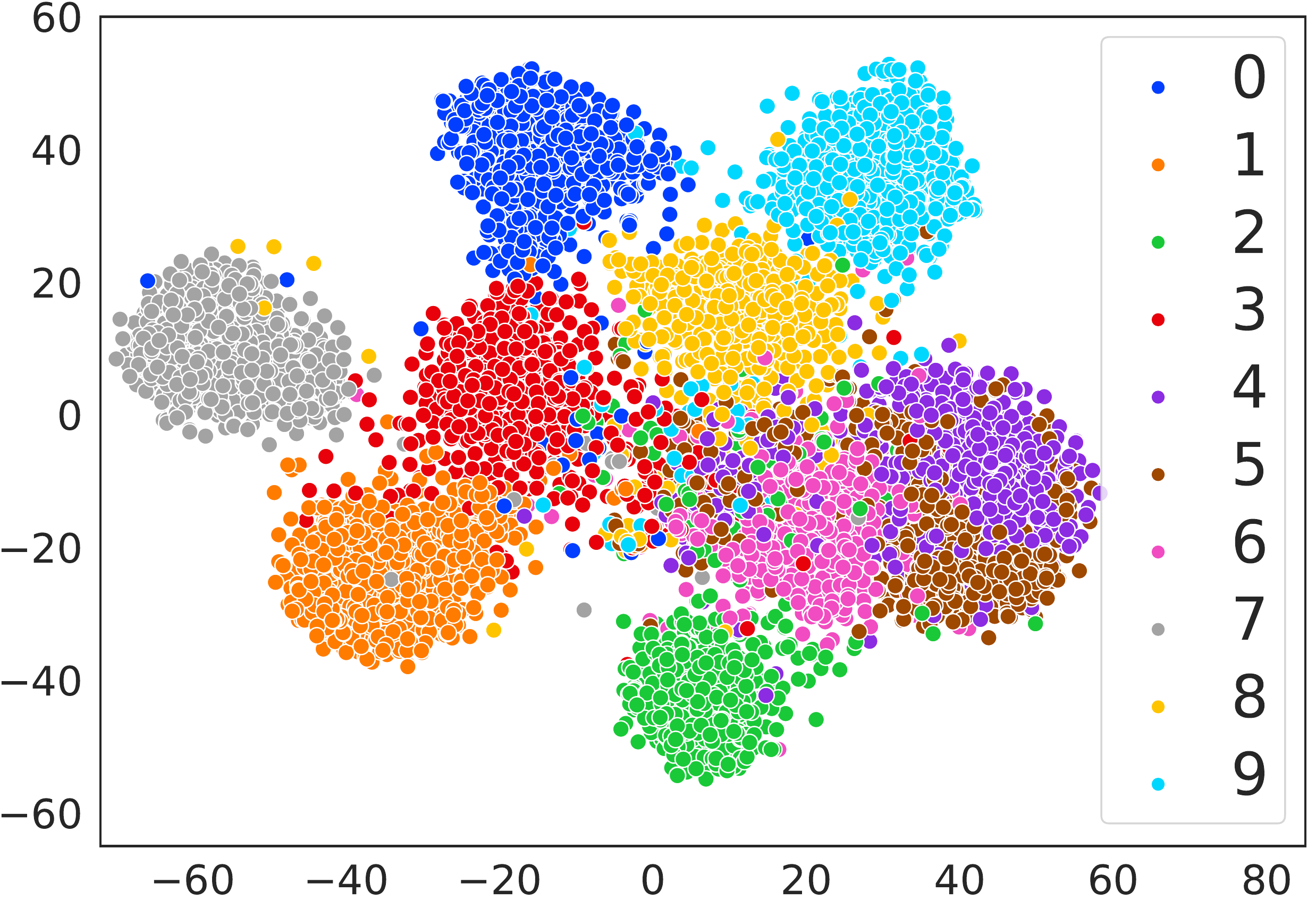}  
\end{subfigure}
\caption{\textbf{Qualitative results on Kinetics-400.} Left: model trained on the labelled data only. Right: our model after end-to-end training on both labelled and unlabelled data. Color denotes the ground truth.}
\label{fig:kinetics_quali}
\end{figure}

We compare the features on videos in~\cref{fig:kinetics_quali}. For visualisation purpose, we randomly choose unlabelled instances from 10 classes from Kinetics-400, we can see that our model can successfully separate novel classes into compact groups, while the novel categories are projected very close to each other for the baseline model trained with full supervision on the labelled data.

\subsection{Ablation study}

\textbf{Component analysis.} We validate the effectiveness of each component of our method during training on both image and video benchmarks. The results are reported in~\cref{tab:ablation}. As can be seen, all components are important to our method. Removing the BCE loss causes the most performance drop. Namely, the ACC drops from $93.4\% \to 32.2\%$ on CIFAR-10, $76.4\% \to 15.5\%$ on CIFAR-100 and $56.5\% \to 5.8\%$ on Kinetics-400, suggesting that the WTA-hashing comparison can indeed generate reliable pairwise pseudo labels for the BCE loss.
Removing the contrastive learning leads to significant performance drop (row 4 vs row 7). 
By comparing rows 5-7, we can see that category (NCE-C) or instance (NCE-I) discrimination alone is not good enough for the task of novel category discovery, and incorporating both is very effective.

\begin{table}[htb]
\centering
\caption{\textbf{Ablation study on image and video benchmarks.} MSE: consistency constraint; CE: cross entropy loss; BCE: binary cross entropy loss; NCE-I: NCE for instance discrimination; NCE-C: NCE for category discrimination.}
\label{tab:ablation}
\centering
\resizebox{0.48\textwidth}{!}{%
\begin{tabular}{l c c c c c c c c}
\toprule
  No & MSE  & CE  & BCE & NCE-I  & NCE-C & CIFAR-10 & CIFAR-100 &  Kinetics-400 \\
\midrule
(1) & \xmark & \cmark & \cmark & \cmark & \cmark & 91.3 $\pm$ 1.4\% & 74.7 $\pm$ 2.9\% & 50.9\% \\
(2) & \cmark  & \xmark & \cmark& \cmark & \cmark & 77.4 $\pm$ 5.4\% & 68.5 $\pm$ 4.7\% & 24.5\% \\
(3) & \cmark  & \cmark & \xmark& \cmark & \cmark & 32.2 $\pm$ 1.8\% & 15.5 $\pm$ 1.2\% & 5.8\% \\
(4) & \cmark  & \cmark & \cmark& \xmark & \xmark & 89.8 $\pm$ 0.9\% & 70.1 $\pm$ 4.5\% & 47.6\% \\
(5) & \cmark  & \cmark & \cmark& \xmark & \cmark & 91.3 $\pm$ 0.8\% & 76.0 $\pm$ 2.4\% & 52.8\% \\
(6) & \cmark  & \cmark & \cmark& \cmark & \xmark & 91.6 $\pm$ 1.2\% & 75.2 $\pm$ 3.1\% & 51.7\% \\
(7) & \cmark  & \cmark & \cmark& \cmark & \cmark & \textbf{93.4 $\pm$ 0.6\%} & \textbf{76.4 $\pm$ 2.8\%} & \textbf{56.5\%} \\
\bottomrule
\end{tabular}
}
\end{table}

\begin{table}[htb]
\centering
\caption{\textbf{Multi-modal contrastive learning.} ``$(a, a)$'' denotes audio-modal contrastive; ``$(v, v)$'' denotes visual-modal contrastive; and ``$(a, v)$'' denotes cross-modal contrastive.}
\label{tab:multi_cl}
\resizebox{0.35\textwidth}{!}{%
\begin{tabular}{l c c c}
\toprule
  No & NCE-I & NCE-C & Kinetics-400  \\
\midrule
   (1) & $(a, a)$ + $(v, v)$ & - & 46.2\% \\
   (2) & $(a, a)$ + $(v, v)$ &  $(a, a)$ + $(v, v)$ & 51.4\% \\
   (3) & $(a, a)$ + $(v, v)$ &  $(a, v)$ & 49.1\% \\
   (4) & $(a, v)$          & - & 51.7\% \\
   (5) & $(a, v)$          &  $(a, a)$ + $(v, v)$  & 56.1\% \\
   (6) & $(a, v)$          &  $(a, v)$  & \textbf{56.5\%} \\
 \bottomrule
\end{tabular}
}
\end{table}
\textbf{Multi-modal contrastive learning.} 
As mentioned in~\cref{sec:method:cl}, for multi-modal data like videos, we can have multiple contrastive options. Namely, we can conduct within-modal  and cross-modal contrastive learning for each of NCE-I and NCE-C. 
We present the results of different options in~\cref{tab:multi_cl}. 
By comparing rows 1 and 4, we can see that cross-modal is more effective than within-modal instance discrimination. It can be seen from rows 1-3 and rows 4-6 that NCE-C can effectively improve the representation for novel category discovery. Also, it is important to consistently use either within-modal or cross-modal discrimination for NCE-I and NCE-C (row 2 vs row 3; row 5 vs row 6). Overall, cross-modal contrastive learning for NCE-I and NCE-C (row 6) performs notably better than all other cases. We hypothesize this is because enforcing the cross-modal agreement can avoid the modal falling into the trivial solution, such as simply identifying the most salient visual or audio features. The modality discrepancy can effectively avoid such cases, thus improving the quality of learned representation.

\subsection{Analysis on WTA}
\noindent \textbf{Hyperparamters of WTA}.
We obtain the two hyperparamters of WTA, i.e. threshold $\mu$ and window size $k$, by examining different values on the labelled data. To do so, we further split the labelled data into a smaller labelled subset and an unlabeled subset (i.e. pretending part of the labelled data to be unlabelled), and find $\mu$ and $k$ that give the best results on the unlabelled subset. 
More specifically, for Kinetics-400, we split 350-class labelled set into a 300-class labelled subset and a 50-class unlabelled subset. For VGG-Sound, we split the 270-class labelled subset into 231-class labelled subset and 39-class unlabelled subset respectively. The number of unlabelled classes used here are the same as the unlabelled classes in the main experiments.
Taking the empirical WTA window size  $k=4$ in~\cite{Yagnik2011thepower}, we experiment with different threshold $\mu$ and report the results in~\cref{tab:sweep_thres}. We find that the performance is generally stable for $\mu$ greater than 200. We set $\mu$ to 240 in our experiments. Note that the feature vectors have a dimension of 512. 
 \begin{table}[htb]
 \centering
 \caption{\textbf{Performance of different WTA threshold.}}
 \label{tab:sweep_thres}
 \resizebox{0.4\textwidth}{!}{
 \begin{tabular}{l l c c c c c}
 \toprule
   Dataset & 130 & 180 & 200 & 240 & 260 & 300 \\
 \midrule
   Kinetics-400 & 19.4 & 39.7 & 55.2 & 54.5 & 54.8  & 54.5 \\
 \midrule
   VGG-Sound & 21.5 & 41.6 & 49.6 & 51.3 & 50.9 & 47.9 \\
 \bottomrule
 \end{tabular}
 }
 \end{table}
Given $\mu=240$, we sweep different $k$ and report the results in~\cref{tab:sweep_widow}. We find that the $k=4$ and $k=8$ perform comparably well, and they are both better than $k=2$ and $k=16$. Hence, we simply use the $k=4$ in our experiments.
 \begin{table}[htb]
 \centering
 \caption{\textbf{Performance of different WTA window size.}}
 \label{tab:sweep_widow}
 \resizebox{0.4\textwidth}{!}{
 \begin{tabular}{l l c c c c}
 \toprule
   Dataset & $k=2$ & $k=4$ & $k=8$ & $k=16$ \\
 \midrule
   Kinetics-400 & 52.6 & 56.7 & 55.9 & 49.4 \\
 \midrule
   VGG-Sound & 46.2 & 51.8 & 51.0 & 48.3 \\
 \bottomrule
 \end{tabular}
 }
 \end{table}

\noindent \textbf{Comparing WTA with other alternatives}.
To better understand the role of WTA in our framework, we investigate other alternatives, namely, cosine similarity, ranking statistics, and nearest-neighbour, to transfer pseudo labels on-the-fly for both~\cite{han20automatically} and our framework. We report the results in \cref{tab:wta_other}. It can be seen from the last row  that simply replacing ranking statistics by WTA for~\cite{han20automatically} only shows marginal improvement, while our method significantly outperforms~\cite{han20automatically} in all cases (using WTA or any other alternative). WTA consistently performs better than ranking statistics.
Intuitively, global ranking (ranking statistics) focuses on the most significant values in the feature space, while WTA considers multiple partial orders in random subsets of feature dimensions. Thus, WTA will consider values spread out the feature space when making the decision, avoiding the comparison to be dominated by high frequency noise or biased by small local regions, which is likely to be the case for global ranking. For example, as shown in~\cite{zhou20lookinto}, modern CNNs are likely to focus only on the (local) parts when recognizing objects (e.g., the beak of a bird), while sometimes the model could focus on the wrong parts, resulting completely wrong predictions. By considering multiple partial rankings, WTA can have a holistic view of the object, leading to more reliable comparison results in a unified framework learning.

\begin{table}[htb]
  \centering
  \caption{\textbf{WTA vs other alternatives.} We compare results of ~\cite{han20automatically}/our method using different pairwise pseudo label generation methods. }
  \label{tab:wta_other}
  \resizebox{0.45\textwidth}{!}{
  \begin{tabular}{l c c c}
  \toprule
    Method & CIFAR-10 & CIFAR-100 & Kinetics-400\\
  \midrule
    cosine & 90.1\%/92.1\% & 73.3\%/75.4\% &  28.3\%/35.4\%\\
  \midrule
    nearest-neighbour & 90.2\%/91.8\% &  69.7\%/73.6\% &  22.7\%/19.6\%\\
  \midrule
    ranking statistics~\cite{han20automatically} & 90.4\%/92.3\% & 73.2\%/74.7\% &  31.2\%/37.4\% \\
  \midrule
    WTA & 90.5\%/\textbf{93.4\%} & 73.4\%/\textbf{76.4\%} &  38.5\%/\textbf{56.5\%} \\
  \bottomrule
  \end{tabular}
  }
  \end{table}
\section{Conclusion}
We have presented a flexible end-to-end framework to tackle the challenging problem of novel category discovery. First, we extended the conventional contrastive learning to perform instance discrimination as well as category discrimination jointly by making full use of the labelled data and unlabelled data. 
Second, to successfully transfer knowledge from the labelled data to the unlabelled data, we employed the WTA hashing algorithm to generate pair-wise weak pseudo labels for training on unlabelled data, which is the key to automatically partition the unlabelled data into proper groups after training. Third, for multi-modal data, we investigated different ways of contrastive learning and we empirically found that cross-modal noise contrastive estimation performs consistently better than other options. Last, we thoroughly evaluated our approach on challenging image and video benchmarks and obtain superior results in all cases.

{\small
\bibliographystyle{ieee_fullname}
\bibliography{novel_video}

\begin{thebibliography}{10}\itemsep=-1pt

\bibitem{alayrac20selfsup}
Jean-Baptiste Alayrac, Adrià Recasens, Rosalia Schneider, Relja Arandjelović,
  Jason Ramapuram, Jeffrey~De Fauw, Lucas Smaira, Sander Dieleman, and Andrew
  Zisserman.
\newblock Self-supervised multimodal versatile networks.
\newblock {\em arXiv prepreint arXiv:2006.16228}, 2020.

\bibitem{alwassel2019self}
Humam Alwassel, Dhruv Mahajan, Lorenzo Torresani, Bernard Ghanem, and Du Tran.
\newblock Self-supervised learning by cross-modal audio-video clustering.
\newblock {\em arXiv preprint arXiv:1911.12667}, 2019.

\bibitem{asano2020labelling}
Yuki~M. Asano, Mandela Patrick, Christian Rupprecht, and Andrea Vedaldi.
\newblock Labelling unlabelled videos from scratch with multi-modal
  self-supervision.
\newblock {\em arXiv prepreint arXiv:2006.13662}, 2020.

\bibitem{chapelle2006semi}
Olivier Chapelle, Bernhard Scholkopf, and Alexander Zien.
\newblock {\em Semi-Supervised Learning}.
\newblock MIT Press, 2006.

\bibitem{chen2020vggsound}
Honglie Chen, Weidi Xie, Andrea Vedaldi, and Andrew Zisserman.
\newblock Vggsound: A large-scale audio-visual dataset.
\newblock In {\em ICASSP}, 2020.

\bibitem{chen2020simple}
Ting Chen, Simon Kornblith, Mohammad Norouzi, and Geoffrey Hinton.
\newblock A simple framework for contrastive learning of visual
  representations.
\newblock {\em ICML}, 2020.

\bibitem{cheng2019panoptic}
Bowen Cheng, Maxwell~D Collins, Yukun Zhu, Ting Liu, Thomas~S Huang, Hartwig
  Adam, and Liang-Chieh Chen.
\newblock Panoptic-deeplab.
\newblock {\em arXiv preprint arXiv:1910.04751}, 2019.

\bibitem{Comaniciu02meanshift}
Dorin Comaniciu and Peter Meer.
\newblock Mean shift: A robust approach toward feature space analysis.
\newblock {\em IEEE TPAMI}, 1979.

\bibitem{deng2009imagenet}
Jia Deng, Wei Dong, Richard Socher, Li-Jia Li, Kai Li, and Li Fei-Fei.
\newblock Imagenet: A large-scale hierarchical image database.
\newblock In {\em CVPR}, 2009.

\bibitem{Dizaji2017deepclustering}
Kamran~Ghasedi Dizaji, Amirhossein Herandi, Cheng Deng, Weidong Cai, and Heng
  Huang.
\newblock Deep clustering via joint convolutional autoencoder embedding and
  relative entropy minimization.
\newblock In {\em ICCV}, 2017.

\bibitem{doersch2015unsupervised}
Carl Doersch, Abhinav Gupta, and Alexei~A. Efros.
\newblock Unsupervised visual representation learning by context prediction.
\newblock In {\em ICCV}, 2015.

\bibitem{Fini_2021_ICCV}
Enrico Fini, Enver Sangineto, Stéphane Lathuilière, Zhun Zhong, Moin Nabi,
  and Elisa Ricci.
\newblock A unified objective for novel class discovery.
\newblock In {\em ICCV}, 2021.

\bibitem{fontanel2020boosting}
Dario Fontanel, Fabio Cermelli, Massimiliano Mancini, Samuel~Rota Bul{\`o},
  Elisa Ricci, and Barbara Caputo.
\newblock Boosting deep open world recognition by clustering.
\newblock {\em arXiv preprint arXiv:2004.13849}, 2020.

\bibitem{gidaris2018unsupervised}
Spyros Gidaris, Praveer Singh, and Nikos Komodakis.
\newblock Unsupervised representation learning by predicting image rotations.
\newblock {\em ICLR}, 2018.

\bibitem{gutmann2010noise}
Michael Gutmann and Aapo Hyv{\"a}rinen.
\newblock Noise-contrastive estimation: A new estimation principle for
  unnormalized statistical models.
\newblock In {\em Proceedings of the Thirteenth International Conference on
  Artificial Intelligence and Statistics}, pages 297--304, 2010.

\bibitem{han20automatically}
Kai Han, Sylvestre-Alvise Rebuffi, Sebastien Ehrhardt, Andrea Vedaldi, and
  Andrew Zisserman.
\newblock Automatically discovering and learning new visual categories with
  ranking statistics.
\newblock In {\em ICLR}, 2020.

\bibitem{han21autonovel}
Kai Han, Sylvestre-Alvise Rebuffi, Sebastien Ehrhardt, Andrea Vedaldi, and
  Andrew Zisserman.
\newblock Autonovel: Automatically discovering and learning novel visual
  categories.
\newblock {\em IEEE TPAMI}, 2021.

\bibitem{han2019learning}
Kai Han, Andrea Vedaldi, and Andrew Zisserman.
\newblock Learning to discover novel visual categories via deep transfer
  clustering.
\newblock In {\em ICCV}, 2019.

\bibitem{he2020moco}
Kaiming He, Haoqi Fan, Yuxin Wu, Saining Xie, and Ross Girshick.
\newblock Momentum contrast for unsupervised visual representation learning.
\newblock {\em CVPR}, 2020.

\bibitem{he2016deep}
Kaiming He, Xiangyu Zhang, Shaoqing Ren, and Jian Sun.
\newblock Deep residual learning for image recognition.
\newblock In {\em CVPR}, 2016.

\bibitem{Hsu18_L2C}
Yen-Chang Hsu, Zhaoyang Lv, and Zsolt Kira.
\newblock Learning to cluster in order to transfer across domains and tasks.
\newblock In {\em ICLR}, 2018.

\bibitem{Hsu18_CCL}
Yen-Chang Hsu, Zhaoyang Lv, Joel Schlosser, Phillip Odom, and Zsolt Kira.
\newblock A probabilistic constrained clustering for transfer learning and
  image category discovery.
\newblock In {\em CVPR Deep-Vision workshop}, 2018.

\bibitem{Hsu19_MCL}
Yen-Chang Hsu, Zhaoyang Lv, Joel Schlosser, Phillip Odom, and Zsolt Kira.
\newblock Multi-class classification without multi-class labels.
\newblock In {\em ICLR}, 2019.

\bibitem{ji20123d}
Shuiwang Ji, Wei Xu, Ming Yang, and Kai Yu.
\newblock 3d convolutional neural networks for human action recognition.
\newblock {\em IEEE TPAMI}, 2012.

\bibitem{ji2019invariant}
Xu Ji, Jo{\~a}o~F Henriques, and Andrea Vedaldi.
\newblock Invariant information clustering for unsupervised image
  classification and segmentation.
\newblock In {\em ICCV}, pages 9865--9874, 2019.

\bibitem{kay2017kinetics}
Will Kay, Joao Carreira, Karen Simonyan, Brian Zhang, Chloe Hillier, Sudheendra
  Vijayanarasimhan, Fabio Viola, Tim Green, Trevor Back, Paul Natsev, et~al.
\newblock The kinetics human action video dataset.
\newblock {\em arXiv preprint arXiv:1705.06950}, 2017.

\bibitem{khosla2020supervised}
Prannay Khosla, Piotr Teterwak, Chen Wang, Aaron Sarna, Yonglong Tian, Phillip
  Isola, Aaron Maschinot, Ce Liu, and Dilip Krishnan.
\newblock Supervised contrastive learning.
\newblock {\em arXiv preprint arXiv:2004.11362}, 2020.

\bibitem{Krizhevsky09cifar}
Alex Krizhevsky and Geoffrey Hinton.
\newblock Learning multiple layers of features from tiny images.
\newblock {\em Technical report}, 2009.

\bibitem{kuhn1955hungarian}
Harold~W Kuhn.
\newblock The hungarian method for the assignment problem.
\newblock {\em Naval research logistics quarterly}, 1955.

\bibitem{kuznetsova2020open}
Alina Kuznetsova, Hassan Rom, Neil Alldrin, Jasper Uijlings, Ivan Krasin, Jordi
  Pont-Tuset, Shahab Kamali, Stefan Popov, Matteo Malloci, Alexander
  Kolesnikov, et~al.
\newblock The open images dataset v4.
\newblock {\em International Journal of Computer Vision}, pages 1--26, 2020.

\bibitem{laine2016temporal}
Samuli Laine and Timo Aila.
\newblock Temporal ensembling for semi-supervised learning.
\newblock In {\em ICLR}, 2017.

\bibitem{MackQueen67_Kmeans}
James MacQueen.
\newblock Some methods for classification and analysis of multivariate
  observations.
\newblock In {\em Proceedings of the Fifth Berkeley Symposium on Mathematical
  Statistics and Probability}, 1967.

\bibitem{morgado2020avid}
Pedro Morgado, Nuno Vasconcelos, and Ishan Misra.
\newblock Audio-visual instance discrimination with cross-modal agreement.
\newblock {\em arXiv prepreint arXiv:2004.12943}, 2020.

\bibitem{oliver2018realistic}
Avital Oliver, Augustus Odena, Colin Raffel, Ekin~D. Cubuk, and Ian~J.
  Goodfellow.
\newblock Realistic evaluation of deep semi-supervised learning algorithms.
\newblock In {\em NeurIPS}, 2018.

\bibitem{park2019specaugment}
Daniel~S Park, William Chan, Yu Zhang, Chung-Cheng Chiu, Barret Zoph, Ekin~D
  Cubuk, and Quoc~V Le.
\newblock Specaugment: A simple data augmentation method for automatic speech
  recognition.
\newblock {\em arXiv preprint arXiv:1904.08779}, 2019.

\bibitem{patrick2020multi}
Mandela Patrick, Yuki~M Asano, Ruth Fong, Jo{\~a}o~F Henriques, Geoffrey Zweig,
  and Andrea Vedaldi.
\newblock Multi-modal self-supervision from generalized data transformations.
\newblock {\em arXiv preprint arXiv:2003.04298}, 2020.

\bibitem{rebuffi20lsdc}
Sylvestre-Alvise Rebuffi, Sebastien Ehrhardt, Kai Han, Andrea Vedaldi, and
  Andrew Zisserman.
\newblock Lsd-c: Linearly separable deep clusters.
\newblock {\em arXiv preprint arXiv:2006.10039}, 2020.

\bibitem{rebuffi20SSL}
Sylvestre-Alvise Rebuffi, Sebastien Ehrhardt, Kai Han, Andrea Vedaldi, and
  Andrew Zisserman.
\newblock Semi-supervised learning with scarce annotations.
\newblock In {\em CVPR Deep Vision Workshop}, 2020.

\bibitem{sutskever2013importance}
Ilya Sutskever, James Martens, George Dahl, and Geoffrey Hinton.
\newblock On the importance of initialization and momentum in deep learning.
\newblock In {\em ICML}, 2013.

\bibitem{szegedy2015going}
Christian Szegedy, Wei Liu, Yangqing Jia, Pierre Sermanet, Scott Reed, Dragomir
  Anguelov, Dumitru Erhan, Vincent Vanhoucke, and Andrew Rabinovich.
\newblock Going deeper with convolutions.
\newblock In {\em CVPR}, 2015.

\bibitem{tan2020efficientdet}
Mingxing Tan, Ruoming Pang, and Quoc~V Le.
\newblock Efficientdet: Scalable and efficient object detection.
\newblock In {\em CVPR}, 2020.

\bibitem{tarvainen2017mean}
Antti Tarvainen and Harri Valpola.
\newblock Mean teachers are better role models: Weight-averaged consistency
  targets improve semi-supervised deep learning results.
\newblock In {\em NeurIPS}, 2017.

\bibitem{tian2019contrastive}
Yonglong Tian, Dilip Krishnan, and Phillip Isola.
\newblock Contrastive multiview coding.
\newblock {\em arXiv preprint arXiv:1906.05849}, 2019.

\bibitem{tran2018closer}
Du Tran, Heng Wang, Lorenzo Torresani, Jamie Ray, Yann LeCun, and Manohar
  Paluri.
\newblock A closer look at spatiotemporal convolutions for action recognition.
\newblock In {\em CVPR}, 2018.

\bibitem{Oord2018RepresentationLW}
Aaron van~den Oord, Yazhe Li, and Oriol Vinyals.
\newblock Representation learning with contrastive predictive coding.
\newblock {\em arXiv prepreint arXiv:1807.03748}, 2018.

\bibitem{wu2018unsupervised}
Zhirong Wu, Yuanjun Xiong, Stella~X Yu, and Dahua Lin.
\newblock Unsupervised feature learning via non-parametric instance
  discrimination.
\newblock In {\em CVPR}, 2018.

\bibitem{Xie16_DEC}
Junyuan Xie, Ross Girshick, and Ali Farhadi.
\newblock Unsupervised deep embedding for clustering analysis.
\newblock In {\em ICML}, 2016.

\bibitem{Yagnik2011thepower}
Jay Yagnik, Dennis Strelow, David~A. Ross, and Ruei sung Lin.
\newblock The power of comparative reasoning.
\newblock In {\em ICCV}, 2011.

\bibitem{zhang2017split}
Richard Zhang, Phillip Isola, and Alexei~A Efros.
\newblock Split-brain autoencoders: Unsupervised learning by cross-channel
  prediction.
\newblock In {\em CVPR}, 2017.

\bibitem{zhao21novel}
Bingchen Zhao and Kai Han.
\newblock Novel visual category discovery with dual ranking statistics and
  mutual knowledge distillation.
\newblock {\em arXiv preprint arXiv:2107.03358}, 2021.

\bibitem{Zhong_2021_CVPR}
Zhun Zhong, Enrico Fini, Subhankar Roy, Zhiming Luo, Elisa Ricci, and Nicu
  Sebe.
\newblock Neighborhood contrastive learning for novel class discovery.
\newblock In {\em CVPR}, 2021.

\bibitem{zhong2021openmix}
Zhun Zhong, Linchao Zhu, Zhiming Luo, Shaozi Li, Yi Yang, and Nicu Sebe.
\newblock Openmix: Reviving known knowledge for discovering novel visual
  categories in an open world.
\newblock In {\em CVPR}, 2021.

\bibitem{zhou20lookinto}
Mohan Zhou, Yalong Bai, Wei Zhang, Tiejun Zhao, and Tao Mei.
\newblock Look-into-object: Self-supervised structure modeling for object
  recognition.
\newblock In {\em CVPR}, 2020.

\bibitem{zhou2019objects}
Xingyi Zhou, Dequan Wang, and Philipp Kr{\"a}henb{\"u}hl.
\newblock Objects as points.
\newblock {\em arXiv preprint arXiv:1904.07850}, 2019.

\end{thebibliography}
}

\clearpage
\appendix\section*{Appendices}
\section{Effects of different WTA hyperparameters on test set}
To further validate the effectiveness of the WTA hyperparameters tuning method (described in section 4.4 of the main paper) under our setting, we conduct experiments on real test set. More specifically, we first take the empirical WTA window size  $k=4$ in~\cite{Yagnik2011thepower} to run experiment with different threshold $\mu$ and report the results in~\cref{tab:sweep_thres_on_test}. We find that conclusion is generally consistent with what we observe in ``validation set'' (i.e., the subset in the labelled data that is pretended to be unlabelled), e.g.,  the performance is generally stable for $\mu$ greater than 200. For kinetics-400, the best performance (55.2) is achieved when $\mu = 200$ in ``validation set'', whereas the best number (56.5) is observed when $\mu = 240$ in test set. For VGG-Sound, the best performance (51.3) is obtained when $\mu = 240$ in ``validation set'', whereas the best number (50.2) is observed when $\mu=200$ in test set. Given that the performance difference between $\mu=240$ and $\mu=200$ is small, and they are neighbouring values in the sweeping set, the hyperparemeter tuning method described in section 4.4 of the main paper appears to be an effective method. We choose $\mu = 240$ for both datasets according to the ``validation set'' to slightly favor the performance on VGG-Sound, as the performance on VGG-Sound is generally worse than that on kinetics-400.

 \begin{table}[htb]
 \centering
 \caption{\textbf{Performance of different WTA threshold.}}
 \label{tab:sweep_thres_on_test}
 \resizebox{0.4\textwidth}{!}{
 \begin{tabular}{l l c c c c c}
 \toprule
   Dataset & 130 & 180 & 200 & 240 & 260 & 300 \\
 \midrule
   Kinetcis-400 & 22.7 & 40.8 & 55.3 & 56.5 & 56.2  & 55.8 \\
 \midrule
   VGG-Sound & 21.2 & 44.2 & 50.2 & 50.0 & 49.4 & 49.3 \\
 \bottomrule
 \end{tabular}
 }
 \end{table}

Similarly, given $\mu=240$, we sweep different $k$ and report the results in~\cref{tab:sweep_widow_on_test}. We also find that the $k=4$ and $k=8$ perform comparably well, and they are both better than $k=2$ and $k=16$. The conclusion remains the same as what we find on ``validation set''.
 \begin{table}[htb]
 \centering
 \caption{\textbf{Performance of different WTA window size.}}
 \label{tab:sweep_widow_on_test}
 \resizebox{0.4\textwidth}{!}{
 \begin{tabular}{l l c c c c}
 \toprule
   Dataset & $k=2$ & $k=4$ & $k=8$ & $k=16$ \\
 \midrule
   Kinetcis-400 & 53.4 & 56.5 & 56.1 & 51.2 \\
 \midrule
   VGG-Sound & 49.2 & 50.0 & 51.1 & 49.7 \\
 \bottomrule
 \end{tabular}
 }
 \end{table}

\section{Unknown class number in unlabelled data} 
Following Han et al.~\cite{han20automatically}, we assume the number of the classes, $C^u$
, in the unlabelled data is known a-priori. When $C^u$ is not known, we can use the method introduced in DTC Han et al.~\cite{han2019learning} to estimate $C^u$ first, and then substitute the estimated number into our framework. We evaluate the performance of our approach on ImageNet using the unknown category numbers estimated by DTC.
The estimates are 34/32/31 and the ground-truth numbers are 30/30/30 on the three unlabelled subsets. The average accuracy over three subsets is $84.1\%$ which outperforms Han et al.~\cite{han20automatically} by $3.6\%$.

\section{Unsupervised clustering}
We further experiment with our approach for pure unsupervised clustering on the unlabelled subset of CIFAR10 and CIFAR100, which contains 5 and 20 classes respectively, by simply dropping the labelled data. Our method achieves $84.6\%$ and $61.5\%$  on the two datasets respectively, while the
results by $k$-means baseline (using features extracted by the model trained on the labelled subset) are
$65.5\%$ and $56.6\%$ respectively (see table 1 in the main paper), showing the superiority of our approach. This reveals
that our method is also an effective clustering method.

\end{document}